\documentclass[11pt,a4paper]{article}
\usepackage[hyperref]{emnlp2020}
\usepackage{times}
\usepackage{latexsym}

\usepackage{caption}
\usepackage{url}
\usepackage{algorithm}
\usepackage{algorithmic}
\usepackage{graphicx}
\usepackage{amsmath}
\usepackage{amsfonts}
\usepackage{multirow}
\usepackage{booktabs}
\usepackage{xcolor}
\usepackage{mathrsfs}
\usepackage{arydshln}
\usepackage{multicol}
\usepackage{multirow}
\usepackage{microtype}

\aclfinalcopy 

\title{Adversarial Attack and Defense of Structured Prediction Models}

\author{Wenjuan Han$^1$\footnotemark[1], Liwen Zhang$^{2,3,4}$\footnotemark[1], Yong Jiang$^5$, Kewei Tu$^{2}$\footnotemark[2]\\
$^1$School of Computing, National University of Singapore, Singapore \\
$^2$School of Information Science and Technology, ShanghaiTech University \\
$^3$Shanghai Institute of Microsystem and Information Technology, Chinese Academy of Sciences\\
$^4$University of Chinese Academy of Sciences\\
$^5$Alibaba DAMO Academy, Alibaba Group \\

{\tt dcshanw@nus.edu.sg} \hspace{2em} \\
{\tt \{zhanglw1, tukw\}@shanghaitech.edu.cn} \\
{\tt yongjiang.jy@alibaba-inc.com} \\

}

\date{}

\begin{document}
\maketitle

\renewcommand{\thefootnote}{\fnsymbol{footnote}}
\footnotetext[1]{Equal contributions.}
\footnotetext[2]{Corresponding author.}
\renewcommand{\thefootnote}{\arabic{footnote}}

\begin{abstract}
Building an effective adversarial attacker and elaborating on countermeasures for adversarial attacks for natural language processing (NLP) have attracted a lot of research in recent years. 
However, most of the existing approaches focus on classification problems. In this paper, we investigate attacks and defenses for structured prediction tasks in NLP. 
Besides the difficulty of perturbing discrete words and the sentence fluency problem faced by attackers in any NLP tasks, there is a specific challenge to attackers of structured prediction models: the structured output of structured prediction models is sensitive to small perturbations in the input. 
To address these problems, we propose a novel and unified framework that learns to attack a structured prediction model using a sequence-to-sequence model with feedbacks from multiple reference models of the same structured prediction task. Based on the proposed attack, we further reinforce the victim model with adversarial training, making its prediction more robust and accurate. We evaluate the proposed framework in dependency parsing and part-of-speech tagging. Automatic and human evaluations show that our proposed framework succeeds in both attacking state-of-the-art structured prediction models and boosting them with adversarial training.
\end{abstract}
\section{Introduction}\label{sec:introduction}
Adversarial examples, which contain perturbations to the input of a model that elicit large changes in the output, have been shown to be an effective way of assessing the robustness of models in natural language processing (NLP) \citep{jia2017adversarial,belinkov2017synthetic,hosseini2017deceiving,samanta2017towards,alzantot2018generating,ebrahimi2018hotflip,michel2019evaluation,wang2019improving}. Adversarial training, in which models are trained on adversarial examples, has also been shown to improve the accuracy and robustness of NLP models \cite{goodfellow2014explaining,tramer2017ensemble,yasunaga2018robust}.
So far, most existing methods of generating adversarial examples only work for classification tasks \citep{jia2017adversarial,wang2019improving} and are not designed for structured prediction tasks. 
However, since many structured prediction tasks such as part-of-speech (POS) tagging and dependency parsing are essential building blocks of many AI systems, it is important to study adversarial attack (generating adversarial examples) and defense (adversarial training) of structured prediction models. 


\begin{figure}
\begin{center}
\includegraphics[width=1.0\columnwidth,trim=0 0 0 0,clip]{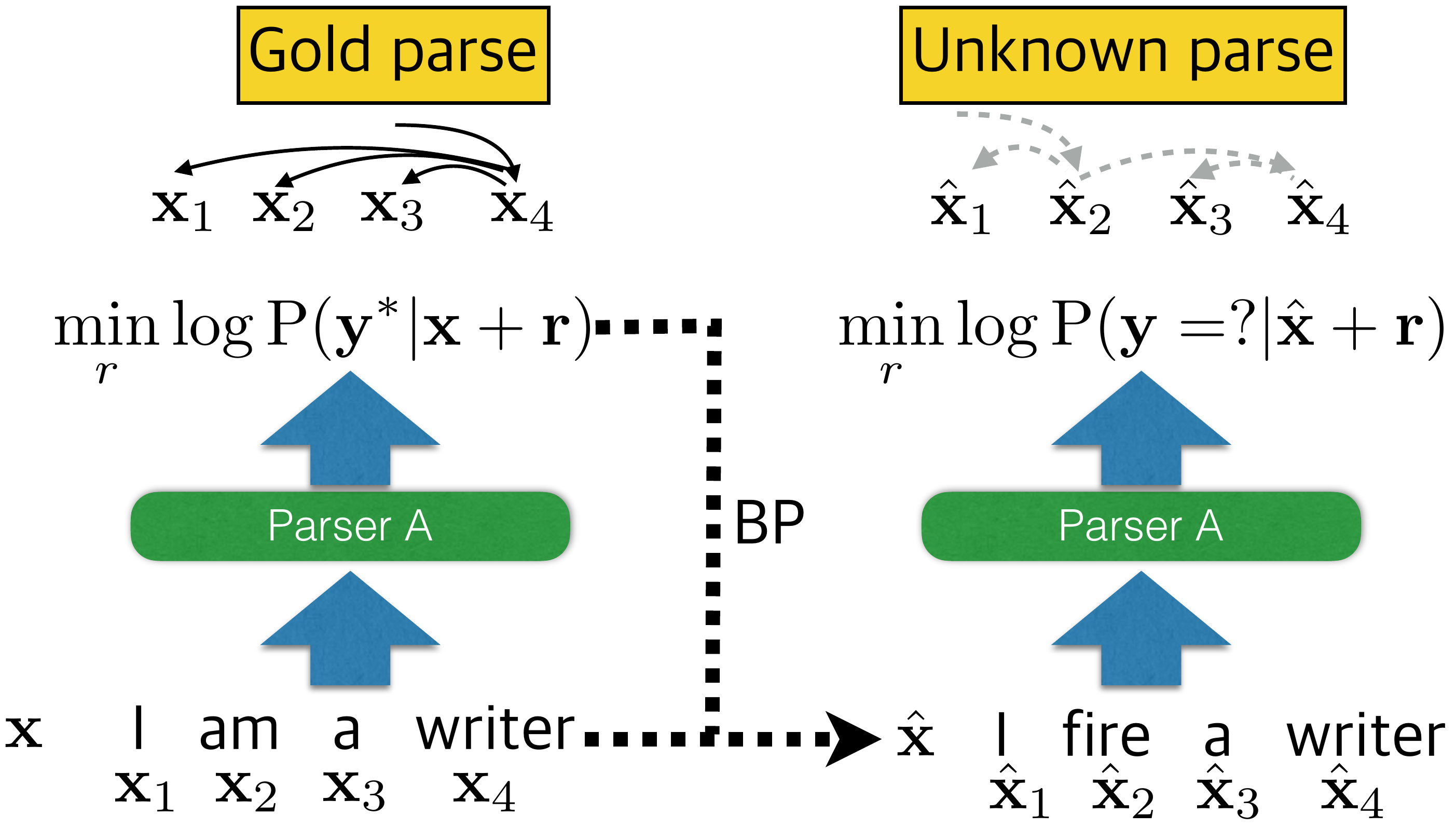}
\caption{An example showing the challenges in attacking a dependency parser using gradient-based methods. A small perturbation to the sentence $x$ changes one word from ``am'' to ``fires''. This change makes the perturbed example \textit{I fires a writer} ungrammatical. Even if the perturbed example is ``I fire a writer'' that meets the rules of grammar, the true output structure is still different from the input sentence ``I am a writer''. More importantly, this true parse is unknown to the attacker, which hinders the next update step.}
\label{pic:baseline}
\vspace{-0.5cm}
\end{center}
\end{figure}

There are multiple challenges that have to be addressed in building an efficient and effective attacker for structured prediction models in NLP. 
\citet{zhang2019generating} pointed out two major problems encountered by attackers of NLP tasks. 
First, since words are discrete, making small disturbances to words in the gradient direction is difficult. 
Secondly, there is no guarantee that the generated adversarial examples are fluent.
In addition to these two problems, there is a unique challenge faced by attackers of structured prediction tasks. 
While small perturbations to images or even texts typically do not change their classification labels, 
small perturbations to sentences in structured prediction may very likely change the true output structures. 
In other words, many structured prediction tasks are very sensitive to small perturbations in the input sentence. 
Consequently, almost all the existing attacking methods are not directly applicable to structured prediction.
To illustrate this challenge, we take adversarial attack of dependency parsing as an example (Figure \ref{pic:baseline}). 
We use the fast gradient sign method (FGSM) \citep{goodfellow2014explaining} as the attack method, which is a classic gradient-based attacker that perturbs the input by minimizing the likelihood of the true output. 
When applied to NLP tasks \citep{miyato2016adversarial}, FGSM perturbs the embeddings of the words in the input sentence and then replaces individual words based on the new embeddings. 
However, there is no guarantee that the new sentence has the same parse tree as the original sentence. Once the true output parse tree becomes unknown, subsequent updates become impossible in FGSM, resulting in perturbation that might be insufficiently adversarial. 
In Figure \ref{pic:baseline}, after just one step of perturbation, the sentence indeed has a different parse tree that is unknown to the attacker.

To address the aforementioned problems, we propose to attack structured prediction models with sequence-to-sequence (seq2seq) sentence generators. 
Before attack, the seq2seq generator is trained by reinforcement learning based on a novelty designed reward function that evaluates the output of the victim structured prediction model against an ensemble of multiple reference models of the same structured prediction task. During attack, the seq2seq generator is simply applied to input sentences to produce adversarial examples.
Our framework has the following features.
\begin{itemize}

    \item Our proposed attacker is a black-box attacker that does not need to know the internal details of the target model (such as the model structure, the hyper-parameters, the training strategy, the training dataset, and gradients over each layer). This ensures that our framework (including attack and defense) can be applied to almost any structured prediction models. 
    \item In contrast to previous black-box attackers, our attacker is an \emph{online} attacker. Once the seq2seq sentence generator is trained, it can generate adversarial examples directly from original sentences during attacks without any optimization procedure and also without the need to access the victim model. This significantly increases the efficiency of the attack.
    \item Most existing methods perform word or character level manipulations and hence cannot change the sentence length. 
    We use a seq2seq generator to modify the whole sentence without this limitation. 
    \item Our generator can utilize some recent pretrained language models (e.g., BERT \citep{devlin2019bert}, GPT-2 \citep{radford2019language}) to improve quality of adversarial examples. 
    

\end{itemize}
We evaluate our framework on the dependency parsing task and the POS tagging task.
Both automatic and human evaluations show that our method outperforms previous approaches in attacking state-of-the-art structured prediction models as well as boosting these models with adversarial training for better accuracy and robustness.
The code and the trained model can be found at \url{https://github.com/WinnieHAN/structure_adv}.



\section{Background}
\subsection{Structured Prediction}
Structured prediction in NLP aims to predict output variables that are mutually dependent or constrained given an input sentence. 
We represent the training data with $N$ samples as $\mathcal{D} = \{\mathbf{x}^{(j)}, \mathbf{y}^{(j)}: j=1,...,N\}$, where $\mathbf{x}^{(j)}$ is the $j$-th sentence and $\mathbf{y}^{(j)}$ is the corresponding structure. The set of all $\mathbf{x}^{(j)}$ is $\mathcal{X}$.
For each $\mathbf{x}$ with length $n$, it can be written as a sequence of tokens $\{\mathbf{x}_{i}:i=1,...,n\}$. We also define $\mathbf{v}$ to represent the concatenation of all the word vectors in sentence $\mathbf{x}$.

A structured prediction model predicts the output $\mathbf{y}$ given an input sentence $\mathbf{x}$ by maximizing the log conditional probability:

\begin{align*}
\arg \max_{\mathbf{y} \in \mathcal{T}} \log P(\mathbf{y}|\mathbf{x};\Theta)
\end{align*}
where $\mathcal{T}$ is the set of all possible outputs and $\Theta$ is the set of parameters. 

We train the model by minimizing the following loss:
\begin{align*}
   \mathcal{L}(\Theta) = - \frac{1}{N}\sum_{(\mathbf{x}^{(j)},\mathbf{y}^{(j)}) \in \mathcal{D}} \log P(\mathbf{y}^{(j)}|\mathbf{x}^{(j)};\Theta)  
\end{align*}{}

\subsection{Word-level Adversarial Attack}
\citet{goodfellow2014explaining} proposed the fast gradient sign method (FGSM) in the image processing field, which uses the direction of the gradient to update image pixels and generate adversarial examples. 
Then \citet{miyato2016adversarial} applied this approach to add perturbations in the word embedding space, though their approach cannot generate adversarial text examples.  
In order to solve the mapping problem from a modified word vector to a word, word level manipulation is used to replace original words \citep{papernot2016crafting}. 
In addition to the replacement manipulation, \citet{samanta2017towards} introduced two new modification strategies: removal and addition.


\subsection{Word-level Adversarial Attack for Structured Prediction}\label{sec:baseline_model}
The gradient of the negative log likelihood with respect to the input in a structured prediction model can be leveraged to find adversarial examples. 
The original input sentence $\mathbf{x}$ is manipulated by adding or subtracting a small adversarial perturbation $\mathbf{r}$ to the vector $\mathbf{v}$. 
Adding $\mathbf{r}$ in the direction of the gradient means that the sentence is modified to decrease the log likelihood so that the model is less likely to predict the correct output. 
We use $\mathbf{\hat{x}}$ to represent $\mathbf{x}$ with perturbation.

The following formula describes the adversarial example:
\begin{align*}
\mathbf{\hat{x}} = search(\mathbf{x}, \mathbf{r}) = search(\mathbf{v}, \mathbf{r})
\end{align*}
where we use $\mathbf{v}$ to represent the concatenation of all the word vectors in sentence $\mathbf{x}$.
$search$ is a searching approach to find an adversarial example $\mathbf{\hat{x}}$ according to perturbed vector $\mathbf{v}+\mathbf{r}$ and $\mathbf{r}$ is calculated by maximizing the loss function as follows.
\begin{align*}  
\mathbf{r} = \arg\max_{\mathbf{r}, || \mathbf{r}|| \leq \epsilon} \{-\log P(\mathbf{y}|\mathbf{x}+\mathbf{r};\Theta)  \}
\end{align*}
where $\epsilon$ is a hyper-parameter to control the scale of the perturbation. 

It is intractable to exactly solve the problem, so an approximate approach is proposed to compute $\mathbf{r}$ as follows:
\begin{align*}  
&\mathbf{r} = \frac{\epsilon \mathbf{g}}{||\mathbf{g}||}
\\
\mathbf{g} =& sign ( \nabla_{\mathbf{v}} \log P(\mathbf{y}|\mathbf{v};\Theta))
\end{align*}

To generate natural and legible adversarial sentences, we search in the word embedding space and replace the original word with a word that is closest  to the perturbed word vector. 
However, as discussed in section \ref{sec:introduction}, this approach can only generate perturbed examples using one perturbation step for structured prediction. 
Moreover, this model cannot guarantee quality (e.g., fluency) of the generating sentences.


\begin{figure}
\begin{center}
\includegraphics[width=0.45\textwidth]{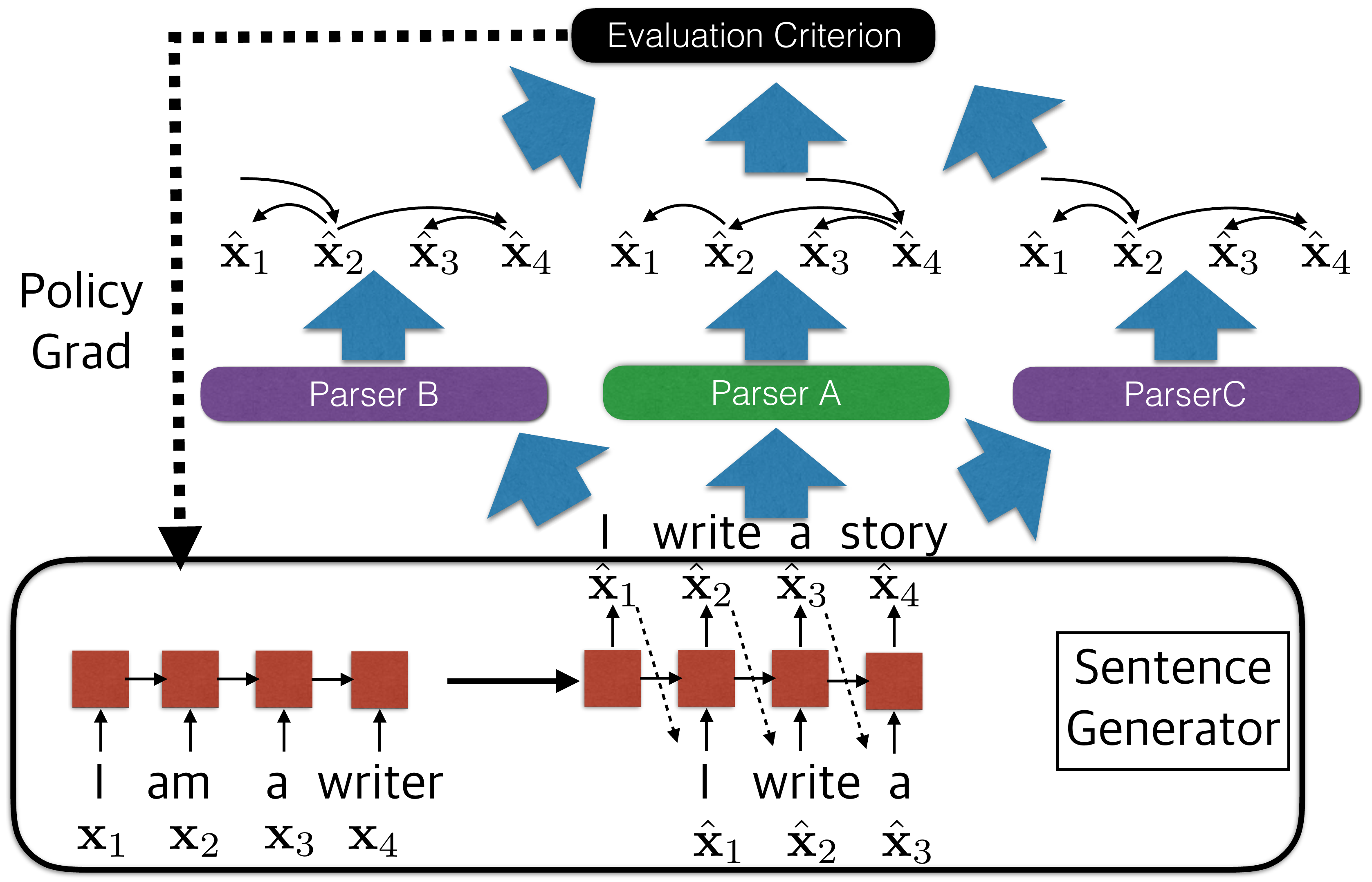}
\caption{Our framework illustrated on the dependency parsing task. 
It consists of three parts: a \textcolor{red}{seq2seq generator}, an evaluation module (including the \textcolor{purple}{reference Parser B}, the  \textcolor{purple}{reference Parser C} and the evaluation criteria), and the \textcolor{green}{victim model A}.} 
\label{pic:framework}
\vspace{-0.5cm}
\end{center}
\end{figure}

\section{Sentence-level Adversarial Attack and Defense}
      
We aim to mislead a structured prediction model by generating adversarial examples $\mathbf{\hat{x}}$ from the original examples $\mathbf{x}$ using a seq2seq generator. 
We train the generator using reinforcement learning following \citet{williams1992simple}. 
The reward function for reinforcement learning evaluates whether the generated sentence could induce an incorrect output from the victim model, and the evaluation is facilitated by two reference models. 
In addition, the reward function also evaluates the quality of the generated sentence.
Figure~\ref{pic:framework} illustrates the overall architecture of our proposed model, which mainly consists of three parts: a generator, an evaluation module, and the victim model. 
We use dependency parsing as our example structured prediction task and name the victim parser as parser A and the reference parsers as Parsers B and C. 



In contrast to the word-level attackers described in section \ref{sec:baseline_model}, our proposed method aims to do sentence-level attacks that are capable of generating a sentence of a different length and structure instead of merely making local word changes.


\subsection{Evaluation Criterion for Structured Outputs} \label{sec:criteria_structure}

Since the structured output is very sensitive to perturbation of the input, the parse of the original sentence $\mathbf{x}$ cannot be treated as the ground truth of the generated adversarial sentence $\mathbf{\hat{x}}$. 
Without knowing the new ground truth, we would not know if the adversarial sentence can indeed mislead parser A to produce an incorrect prediction. 
Thus we make use of two reference parsers B and C to evaluate the prediction of parser A and help guide the generation of truly adversarial examples. Intuitively, if B and C produce the same parse tree, then it is more likely to be correct and can be used as ground-truth to evaluate parser A.


Given a generated sentence $\mathbf{\hat{x}}$, 
if the predicted parse tree $\mathbf{y}^{A}_{\mathbf{\hat{x}}}$ from parser A is greatly different from the predicted trees $\mathbf{y}^{B}_{\mathbf{\hat{x}}}$ and $\mathbf{y}^{C}_{\mathbf{\hat{x}}}$ from parsers B and C, 
while $\mathbf{y}^{B}_{\mathbf{\hat{x}}}$ and $\mathbf{y}^{C}_{\mathbf{\hat{x}}}$ agree with each other, then we think $\mathbf{\hat{x}}$ is a good adversarial example of parser A. The criterion is defined as follows:
\begin{equation} \label{eq:1}
\begin{split}
s_{p}(\mathbf{\hat{x}}) = -f(\mathbf{y}^{A}_{\mathbf{\hat{x}}}, \mathbf{y}^{B}_{\mathbf{\hat{x}}}) - f(\mathbf{y}^{A}_{\mathbf{\hat{x}}}, \mathbf{y}^{C}_{\mathbf{\hat{x}}})  + f(\mathbf{y}^{B}_{\mathbf{\hat{x}}}, \mathbf{y}^{C}_{\mathbf{\hat{x}}})
\end{split}
\end{equation}
where $f(\mathbf{y}, \mathbf{y}^*)$ is a symmetric function that evaluates the difference between two parse trees $\mathbf{y}$ and $\mathbf{y}^*$. 
A higher value of $s_{p}(\mathbf{\hat{x}})$ means $\mathbf{\hat{x}}$ is more adversarial.

The primary criterion for selecting parsers B and C is their parsing accuracy. 
As we defined in Equation \ref{eq:1}, the consensus prediction of parsers B and C is regarded as ground truth, no matter whether the prediction is actually right or wrong. 
Thus parsers B and C should have high accuracy and also different inductive biases so that they are unlikely to make the same mistake. 
In addition, B and C should not be too similar to parser A, because otherwise the first two terms in Equation \ref{eq:1} would become hard to optimize.

\subsection{Evaluation Criteria for Sentence Quality}\label{sec:metric_quality}
We consider two aspects of the sentence quality as follows:
\begin{itemize}
    \item Fluency: Inspired by \citet{holtzman2018learning,xu2018dp,pang2020towards}, we use perplexity on GPT-2 \citep{radford2019language}, a large Transformer language model trained on massive texts, to evaluate the fluency of the generated sentences. 
    We use the negative perplexity as a reward in the learning process. 
    \begin{equation*} \label{eq:2}
    \begin{split}
    s_{f}(\mathbf{\hat{x}}) = -PPL(\mathbf{\hat{x}})
    \end{split}
    \end{equation*}
    \item Meaning Preservation: Adversarial examples that differ too much from the original sentences are less effective in attacks because humans can easily identify them. We use BERTScore \citep{zhang2019bertscore} as another reward in learning to evaluate the similarity between two sentences at the meaning level. We choose to use BERTScore because it correlates better with human judgments than traditional measures such as BLEU \citep{papineni2002bleu}. 
    \begin{equation*} \label{eq:3}
    \begin{split}
    s_{m}(\mathbf{x}, \mathbf{\hat{x}}) = BERTScore (\mathbf{x}, \mathbf{\hat{x}})
    \end{split}
    \end{equation*}
\end{itemize}{}
By maximizing these criteria, we hope to make the adversarial examples look more like human generated sentences and not differ too much from the original sentences in meaning. 

\subsection{Sentence Generator}\label{sec:sentence-generator}

We propose to use a seq2seq model \citep{wang2016attention} as the adversarial sentence generator, which has been widely used in machine translation, dialogue and many other areas.  
The seq2seq model specifies $P(\mathbf{\hat{x}}|\mathbf{x};\Theta)$, the conditional probability of generating an adversarial sentence $\mathbf{\hat{x}}$ given an input sentence $\mathbf{x}$.
We train the model by reinforcement learning guided by our aforementioned criteria.
The objective function is the expected reward based on the sentences from the training corpus $\mathcal{X}$,
\[J(\Theta) = \sum_{\mathbf{x} \in \mathcal{X}} E_{\mathbf{\hat{x}} \sim P(\mathbf{\hat{x}}|\mathbf{x};\Theta)} s(\mathbf{x},\mathbf{\hat{x}})\]
The reward $s(\mathbf{x},\mathbf{\hat{x}})$ is composed of three parts.
    \begin{equation} \label{eq:4}
    \begin{split}
    s(\mathbf{x},\mathbf{\hat{x}}) = \alpha s_{p}(\mathbf{\hat{x}}) + \beta s_{f}(\mathbf{\hat{x}}) + \gamma s_{m}(\mathbf{x},\mathbf{\hat{x}})
    \end{split}
    \end{equation}
where $\alpha, \beta, \gamma$ are tunable hyper-parameters that control the balance between the three parts.
We optimize the objective function with the REINFORCE algorithm \citep{williams1992simple}. 


To further encourage meaning preservation between $\mathbf{x}$ and $\mathbf{\hat{x}}$, we also pretrain the seq2seq model as a denoising auto-encoder before reinforcement learning. 
Specifically, during pretraining we add noise to the hidden states of encoder and train the decoder to recover the input sentence. 
We employ the masking noise method that masks each word of the input sentence by a fixed probability and trains the denoising autoencoder to fill in these ``blanks'' \citep{vincent2008extracting}.

\subsection{Defense against Adversarial Attack}
Following \citet{goodfellow2014explaining}, we use adversarial training to withstand attacks. More specifically, we enhance the victim model
by injecting adversarial examples into the training data and retraining the model with the mixed data.

\section{Experiments on Dependency Parsing}\label{parsing}
We first perform experiments on dependency parsing, a well-known structured prediction task.

\subsection{Data}\label{parsing:data}
Our model does not need labeled data for training but we need a victim parser and two reference parsers in our experiments.
We learn these parsers on an English dataset: Penn Treebank 3.0 (PTB, \citet{marcus1994penn}).
We also use the same data for training and evaluating our model.


\subsection{Parser Selection}

We choose the Deep Biaffine parser (\citet{dozat2016deep}), one of the state-of-the-art graph-based parsers, as the victim parser A. 
For the reference parsers, we choose two other well-known dependency parsers: 
\begin{itemize}
     \item[-] \textbf{Parser B}: StackPTR from \citet{ma2018stack}
     \item[-] \textbf{Parser C}: BiST from \citet{kiperwasser2016simple}
\end{itemize}

The three parsers are trained with PTB.
All the hyper-parameters of these parsers are the same as reported in their papers. 

\begin{table*}[!t]
\centering
\resizebox{\textwidth}{!}{%
\begin{tabular}{r|c|c|c|c|ccc}
\hline
\multicolumn{1}{c|}{} &
  \textbf{Generation Fluency} &
  \multicolumn{3}{c|}{\textbf{Token level Attacking Success Rate}} &
  \multicolumn{3}{c}{\textbf{Sentence level Attacking Success Rate}} \\ \cline{3-8} 
\multicolumn{1}{c|}{} & \textbf{(Perplexity $\downarrow$)}
   &
  \textit{Parser B} &
  \textit{Parser C} &
  \textit{Parsers B\&C} &
  \multicolumn{1}{c|}{\textit{Parser B}} &
  \multicolumn{1}{c|}{\textit{Parser C}} &
  \textit{Parsers B\&C} \\ \hline
Origin   & \textbf{156.02} & 3.2 & 3.6 & 4.6  & \multicolumn{1}{c|}{34.2} & \multicolumn{1}{c|}{35.3} & 40.7 \\
Baseline & 217.02  & 3.7 & 10.1 & 11.5 & \multicolumn{1}{c|}{37.0} & \multicolumn{1}{c|}{64.1} & 67.4 \\
Ours &
  174.16 & \textbf{13.9} & \textbf{19.2} & \textbf{24.1} & \multicolumn{1}{c|}{\textbf{87.4}} &  \multicolumn{1}{c|}{\textbf{86.5}} & \textbf{89.0} \\ \hline
\end{tabular}
}
\caption{Experimental results on dependency parsing based on automatic evaluation. 
``Origin'' shows the results of original sentences in the PTB test set. Lower perplexity is better.}
\label{tab:parsing-auto}
\vspace{-0.4cm}
\end{table*}

\begin{table}[h]
\centering
\resizebox{0.47\textwidth}{!}{%
\begin{tabular}{r|c|c|c}
\hline
         & \textbf{Generation} & \multicolumn{2}{l}{\textbf{Attacking Success Rate}} \\ \cline{3-4} 
         & \textbf{Fluency} $\uparrow$ & \textit{Token}          & \textit{Sentence}       \\ \hline
Baseline & 3.21                               & 10.8                  & 64    \\ \hline
Ours     & \textbf{3.84}                      & \textbf{18.3}        & \textbf{72}          \\ \hline
\end{tabular}
}
\caption{Experimental results on dependency parsing based on human evaluation. Higher is better.}
\label{tab:parsing-human}
\vspace{-0.2cm}
\end{table}

\subsection{Evaluation Metrics}\label{parsing:eval}

Our goal is to generate fluent sentences that are mispredicted by the victim model.
Thus, we evaluate the adversarial examples produced by our model from 2 aspects: generation fluency and attacking efficiency (6 metrics).

\paragraph{Generation Fluency}
We use the perplexity on GPT-2 to evaluate the fluency of the generated sentences.

\paragraph{Attacking efficiency} We evaluate the attacking success rates at the token level and sentence level.
The token level attacking success rate is the percentage of words in the generated adversarial examples that are assigned the wrong head without considering the labels of the dependence type. It is also known as unlabeled attachment score (UAS). 
Sentence-level attacking success rate is the percentage of mispredicted sentences in the generated adversarial examples. 
Due the lack of golden parse trees of generated sentences,
here we leverage the parses predicted by Parsers B and C as ground truth.
The token level and sentence level each has three metrics: predictions of B as ground truth, predictions of C as ground truth, and consensus predictions of B and C as ground truth (discarding the sentences on which they disagree).

\paragraph{Human evaluation} 
We conduct human evaluation of the fluency and attacking efficiency. All the volunteers have a background of linguistic study and are proficient in English. We further train the volunteers with the annotated English PTB treebank. From the adversarial examples generated by our method, we randomly sample 50 examples. 
During labeling, we ask two of them to label the sentences and the third skilled volunteers to double-check the evaluation results.
For fluency, we ask them to rate the fluency of a sentence by an integer from 1 to 5. 5 indicates a sentence is fluent and has no grammatical errors. 1 indicates a sentence is full of grammatical errors and meaningless. For attacking efficiency, we ask them to manually annotate erroneous dependency edges and calculate the error rate in the same way as in automatic evaluation. The predictions of the Parsers B and C are given for reference.

\subsection{Experimental Setup}\label{parsing:setup}

We take the word-level approach in section \ref{sec:baseline_model} as our baseline, which uses a one-step update. 
Intuitively, this approach maintains the length of sentences and perturbs sentences by word-level replacement. 

For our seq2seq generator, we use an attention-based three layers of BiLSTM with hidden vector dimension 1024.
First, we pretrain the seq2seq generator for $3$ epochs with unlabeled sentences from the PTB training set.
The objective function for pretraining is negative conditional log likelihood. 
Then we train the seq2seq generator using reinforcement learning with hyper-parameter $\alpha=1$, $\beta=0.001$, $\gamma=100$.
Adam \citep{kingma2014adam} is used to optimize the parameters with the learning rate is $2e$-$5$. 
The minibatch size during reinforcement learning is 16.
A detailed description of hyper-parameter settings can be found in Appendix A.

\subsection{Experimental Results}\label{parsing:result}

\begin{figure}
\begin{center}
\includegraphics[width=\columnwidth]{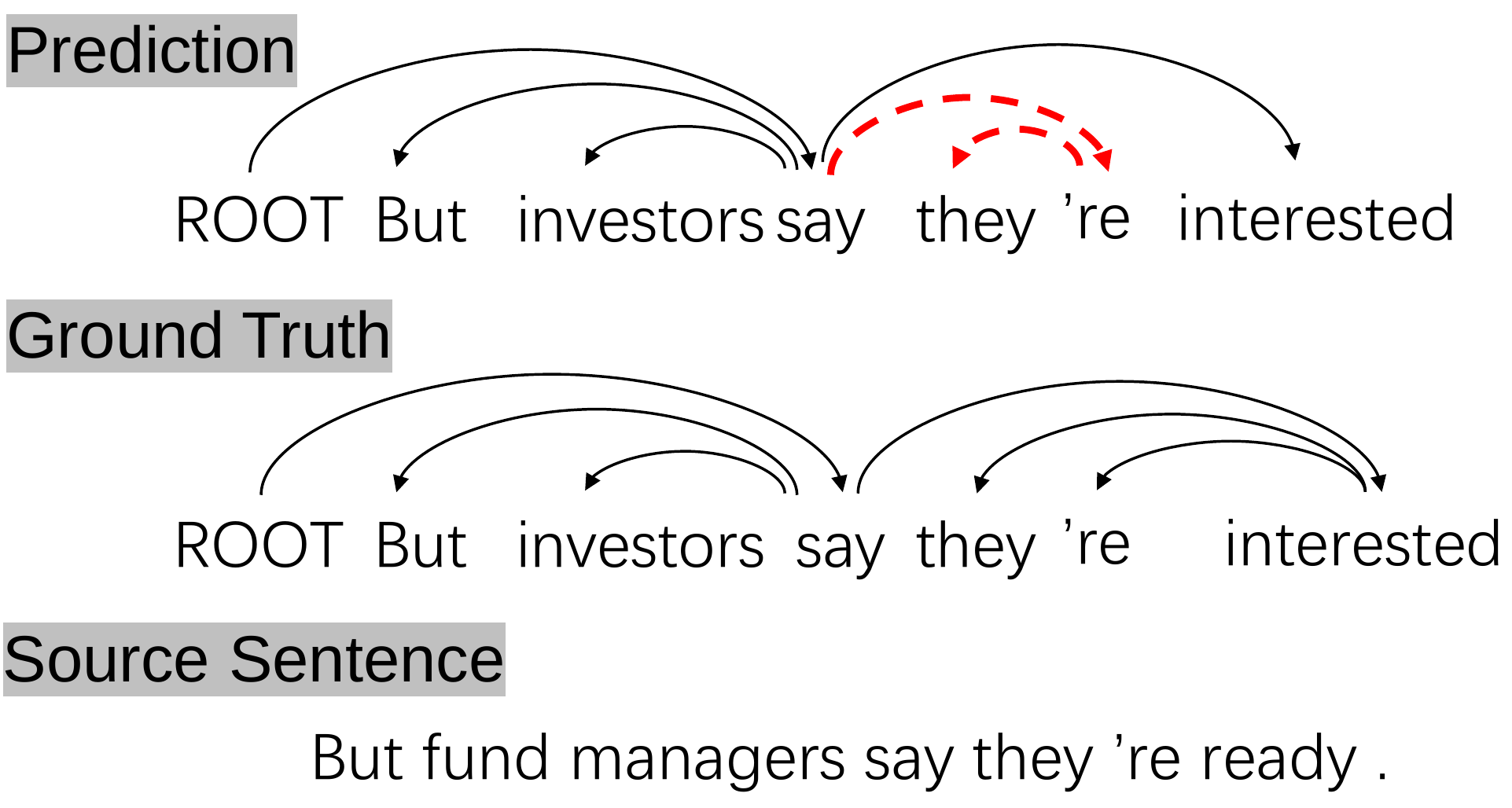}
\caption{Case study of an adversarial example for dependency parsing task. The mispredicted dependencies of victim parser A are highlighted by dotted lines.}
\label{fig:parsing-cases}
\vspace{-0.5cm}
\end{center}
\end{figure}

Table \ref{tab:parsing-auto} shows the automatic evaluation results.
The attacking success rate improvement of our method over the baseline reflects the effectiveness of our reinforcement learning strategy.
Particularly, our method improves the token level and sentence level attacking success rate 13.4\% and 21.6\% on Parsers B\&C, respectively.
It can also be seen that our proposed method maintains good fluency while making successful attacks.
Human evaluation shown in Table \ref{tab:parsing-human} is consistent with automatic evaluation: our proposed method is significantly better than the baseline model at both generation fluency and attacking success rate. For better comparison, we ask volunteers to label the fluency score of the original sentences in PTB and obtain 4.64.
We show an adversarial example in Figure \ref{fig:parsing-cases}.

\begin{table}[]
\centering
\begin{tabular}{c|c|c}
\hline
\multicolumn{2}{c|}{}                                   & \textbf{UAS} \\ \hline
\multicolumn{2}{c|}{\textbf{W/O Adv Train}}                   & 95.42        \\ \hline
\multirow{4}{*}{\textbf{Adv Train}} & \textit{Baseline}     & 95.54   \\ \cline{2-3}                   &  \textit{BLLIP-BC}     & 95.51      \\ \cline{2-3}
                              & \textit{BLLIP-ABC} & 95.46        \\  \cline{2-3}
                              & \textit{Ours}           & \textbf{95.63}         \\ \hline
\end{tabular}%
\caption{Adversarial Training on different datasets for dependency parsing. \textit{Adv Train}: adversarial training. }
\label{tab:dep_adv_train}
\end{table}

\begin{figure}
\begin{center}
\includegraphics[width=\columnwidth]{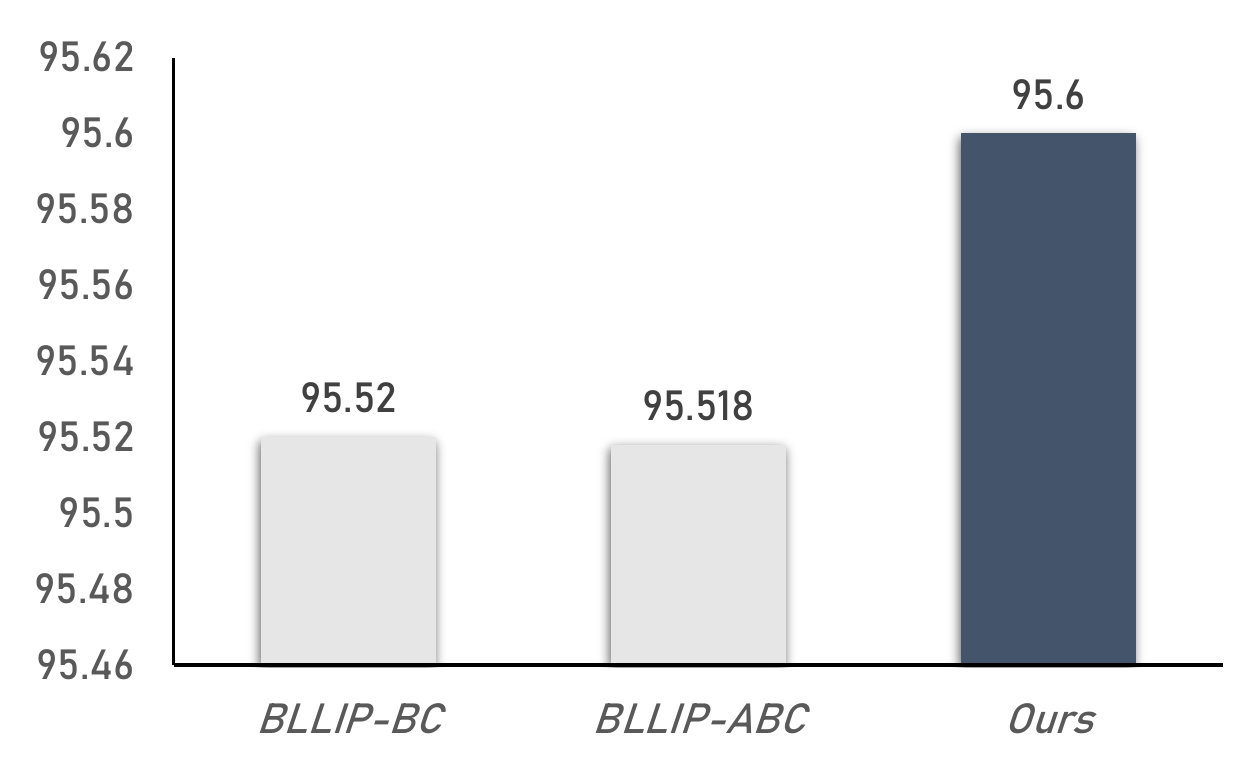}
\caption{Average results of five time retrain using different datasets.}
\label{fig:avg-res}
\vspace{-0.5cm}
\end{center}
\end{figure}

\begin{table*}[h]
\centering
\resizebox{\textwidth}{!}{%
\begin{tabular}{r|c|c|c|c|ccc}
\hline
\multicolumn{1}{c|}{} &
  \textbf{Generation Fluency} &
  \multicolumn{3}{c|}{\textbf{Token level Attacking Success Rate}} &
  \multicolumn{3}{c}{\textbf{Sentence level Attacking Success Rate}} \\ \cline{3-8} 
\multicolumn{1}{c|}{} & \textbf{(Perplexity $\downarrow$)} &
  \textit{Tagger B} &
  \textit{Tagger C} &
  \textit{Tagger B\&C} &
  \multicolumn{1}{c|}{\textit{Tagger B}} &
  \multicolumn{1}{c|}{\textit{Tagger C}} &
  \textit{Tagger B\&C} \\ \hline
Origin   & 156.02 & 1.9 & 2.1 & 3.2  & \multicolumn{1}{c|}{30.6} & \multicolumn{1}{c|}{35.5} & 45.7 \\
Baseline & 354.24  & 3.8 & 4.2 & 6.5 & \multicolumn{1}{c|}{55.6} & \multicolumn{1}{c|}{57.5} & 71.6 \\
Ours &
  \textbf{142.59} & \textbf{9.2} & \textbf{7.3} & \textbf{14.5} & \multicolumn{1}{c|}{\textbf{78.1}} &  \multicolumn{1}{c|}{\textbf{73.3}} & \textbf{89.0} \\ \hline
\end{tabular}
\vspace{-0.4cm}
}
\caption{Experimental results on POS tagging based on automatic evaluation. ``Origin'' shows the results of the original sentences. Lower perplexity is better.}
\label{tab:tagging_auto}
\vspace{-0.4cm}
\end{table*}

\begin{table}[t]
\centering
\resizebox{0.47\textwidth}{!}{%
\begin{tabular}{r|c|c|c}
\hline
         & \textbf{Generation} & \multicolumn{2}{l}{\textbf{Attacking Success Rate}} \\ \cline{3-4} 
         &  \textbf{Fluency} $\uparrow$ & \textit{Token}          & \textit{Sentence}       \\ \hline
Baseline & \textbf{3.98}                               & 1.8                  & 16                   \\ \hline
Ours     & 3.88                      & \textbf{8.1}        & \textbf{52}          \\ \hline
\end{tabular}
}

\caption{Experimental results on POS tagging based on human evaluation. Higher means better.}
\label{tab:tagging_human}
\vspace{-0.4cm}
\end{table}

\subsection{Adversarial Training}
We then conduct experiments on adversarial training and summarize the results in Table \ref{tab:dep_adv_train}. 
We add 2000 adversarial examples to the original training data and retrain the Biaffine parser\footnote{The candidate sentences are generated by the seq2seq generator using sentences in the training dataset as input. Then we drop the sentences that do \textbf{not} meet the criterion: reference parsers B and C predict the same parse trees that are different from the predictions of parser A (namely, the victim parser). Finally, we select the first 2000 sentences from the remaining 2044 sentences as the adversarial examples.}.
We use the predicted parser Tree from Parsers B and C as the ground truth for these adversarial examples.
If the parse trees from Parsers B and C are not the same, we drop the sentence.
In addition to \textit{W/O Adv Train} (result without adversarial training) and \textit{Baseline} (retraining with adversarial examples produced by the word-level approach), we also experiment with the following two baseline methods of collecting 2000 additional training samples the BLLIP dataset\footnote{Brown Laboratory for Linguistic Information Processing (BLLIP) 1987-89 WSJ Corpus Release 1. We choose the BLLIP corpus because it is collected from the same news article source as the WSJ corpus.}: 
\begin{itemize}
    \item[-] \textit{BLLIP-BC}: Sampling sentences on which Parsers B and C predict the same parse trees.
    \item[-] \textit{BLLIP-ABC}: Sampling sentences on which Parsers B and C predict the same parse trees that are different from the predictions of Parser A.
\end{itemize}
We use the predicted parse trees from Parsers B and C as the ground truth for these two kinds of baselines.
It can be seen that adversarial training, with adversarial examples leads to the largest performance gain over the ``no adversarial training'' baseline. 

Although Table~\ref{tab:dep_adv_train} shows that fine-tuning the victim parser A on our adversarial samples achieves better performance, the improvement is small. 
To investigate whether the improvement is significant or not, we retrain the parser A for five times with different random seeds.
We also rerun the \textit{BLLIP-BC} and \textit{BLLIP-ABC} baselines (including the sampling step) for five times with different random seeds.
The learning rate is $5e$-$4$. 
After training for 50 epochs, the average results are shown in Figure~\ref{fig:avg-res}.
It shows that our method outperforms the two baselines.
We also perform Student's t-test:
\begin{itemize}
    \item[-] \textit{BLLIP-BC} and \textit{Ours}: $t$-value is -2.77 and $p$-value is 0.024.
    \item[-] \textit{BLLIP-ABC} and \textit{Ours}: $t$-value is -3.39 and $p$-value is 0.010.
\end{itemize}
Both $p$-values are less than 0.05. That means the advantage of our method is statistically significant.

We also perform human evaluation on the retrained parser. 
The token level attacking success rate drops 1.3 points from 18.3 to 17.0, and the sentence level attacking success rate reduces from 72 to 70.
We perform significance tests on the attacking success rate. The p-value is calculated by using the one-tailed sign test with bootstrap resampling on 50 samples following Chollampatt, Wang, and Ng (2019). We compare the attacking success rate with and without retraining. The p-values (5.42e-20 at the token level and 3.39e-21 at the sentence level) show that the improvement is significant.
 
\section{Experiments on POS Tagging}\label{tagging}

\subsection{Experimental Setup}\label{tagging:setup}
In this section, we apply our method to the part-of-speech tagging task using the tagger from \citet{ma2016end} as the victim model.
For the reference taggers, we choose two state-of-the-art taggers: Stanford POS tagger from \citet{toutanova2003feature} and Senna tagger from \citet{collobert2011natural}. 
All the hyper-parameters of the three taggers are the same as reported in their papers. 
We conduct the experiments on the PTB dataset. 

Similar to dependency parsing, the word level approach in section \ref{sec:baseline_model} is the baseline.
For the adversarial example generator, we use the same structure and pretrain strategy as Section~\ref{parsing:setup}, except that the dimension of hidden state is set to 512.
We train the sentence generator using reinforcement learning with hyper-parameter $\alpha=1$, $\beta=0.001$, $\gamma=30$.
Adam\citep{kingma2014adam} is used to optimize the parameters with learning rate $5e$-$4$. 
The minibatch size during reinforcement learning is 64.
A detailed description of hyper-parameter settings can be found in Appendix B.
We employ the same set of evaluation metrics as in section \ref{parsing:eval}.


\begin{figure}
\begin{center}
\includegraphics[width=\columnwidth]{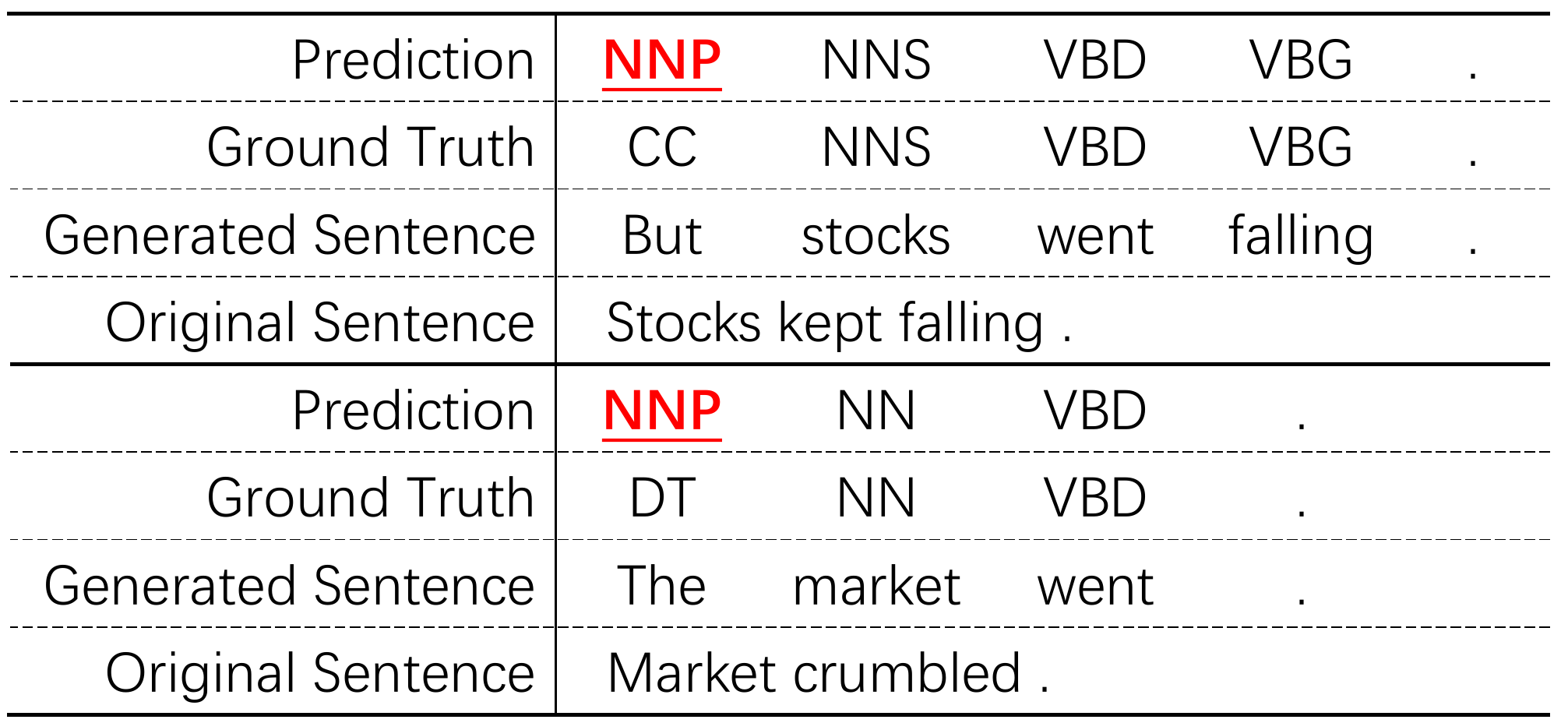}
\caption{Case study of an adversarial example for POS tagging task. The mispredicted POS tags of victim tagger A are highlighted with underlines.}
\label{fig:tagging-cases}
\vspace{-0.5cm}
\end{center}
\end{figure}

\subsection{Experimental Results}\label{tagging:result}
We perform automatic evaluation over all the samples generated from the test dataset.
As shown in Table \ref{tab:tagging_auto}, attacking success rate and fluency of our proposed method are both above those of the baseline, which indicates the effectiveness of our proposed method. 
Particularly, our method improves the token level and sentence level attacking success rate 8.0\% and 17.3\%, respectively.

Similar to the dependency parsing task, Table \ref{tab:tagging_human} shows the result of human evaluation of 50 samples. 
According to human evaluation, the fluency of sentences generated by the two methods is similar, 
but the attacking success rate of our method is significantly higher than the baseline. 
Two example are shown in Figure~\ref{fig:tagging-cases}.

We also conduct experiments on adversarial training with 1000 additional samples produced by our method. 
After retraining, the accuracy of Tagger A improves 0.13 point from 97.55 to 97.68 on PTB the test set. 
Similar to dependency parsing, we perform t-test to measure the statistical significance of the advantage of our method in POS tagging. The resulting p-value is 0.027.

\section{Analysis}

\subsection{Selecting Reference Model}
We mention in the Section~\ref{sec:criteria_structure} that the victim model and the two reference model should differ from each other as much as possible.
In our previous experiments in Section~\ref{parsing} , we use three different types of parsers as the victim parser (Deep Biaffine) and reference parsers (StackPtr and BiST).
Here we investigate the impact of making them similar.
First, we make the two reference parsers similar to the victim parser, by training two Deep Biaffine parsers with different random seeds. We call this AllSame.
Second, we make the two reference parsers similar to each other but different from the victim parser, by training two StackPtr parsers with different random seeds. We call this EvalSame.

Table~\ref{tab:analysis_human} shows that AllSame tends to generate fluent sentences but the sentences are less adversarial. 
This can be explained by the fact that the similarity between the parsers make the first term of Equation~\ref{eq:4} very small and the reward function is dominated by the two sentence quality terms.
EvalSame can be seen to produce slightly higher token level attacking success rate but significantly lower generation fluency. 
Compared with AllSame and EvalSame, our standard method of using two different parsers as the reference models can reach a better attacking success rate, while keeping the sentences relatively fluent.

\subsection{Applicability Analysis}
We repeat our experiment of dependency parsing following the setup of Table \ref{tab:parsing-auto} except for the choice of the victim parser and reference parsers.
We use StackPTR as the victim model while the Deep Biaffine parser and BiST as Parser B and Parser C. Table \ref{tab:parsing-auto-exchange} shows the automatic evaluation results.  
The results show similar trends to those in Table \ref{tab:parsing-auto}, suggesting that our approach is effective to different choices of the victim parser and reference parsers.
\begin{table*}[!t]
\centering
\resizebox{\textwidth}{!}{%
\begin{tabular}{r|c|c|c|c|ccc}
\hline
\multicolumn{1}{c|}{} &
  \textbf{Generation Fluency} &
  \multicolumn{3}{c|}{\textbf{Token level Attacking Success Rate}} &
  \multicolumn{3}{c}{\textbf{Sentence level Attacking Success Rate}} \\ \cline{3-8} 
\multicolumn{1}{c|}{} & \textbf{(Perplexity $\downarrow$)}
   &
  \textit{Parser B} &
  \textit{Parser C} &
  \textit{Parsers B\&C} &
  \multicolumn{1}{c|}{\textit{Parser B}} &
  \multicolumn{1}{c|}{\textit{Parser C}} &
  \textit{Parsers B\&C} \\ \hline
Baseline & 377.36  & 4.5 & 15.9 & 17.5 & \multicolumn{1}{c|}{40.7} & \multicolumn{1}{c|}{74.5} & 74.90 \\
Ours &
  \textbf{244.69} & \textbf{19.6} & \textbf{23.3} & \textbf{26.2} & \multicolumn{1}{c|}{\textbf{70.8}} &  \multicolumn{1}{c|}{\textbf{77.2}} & \textbf{80.1} \\ \hline
\end{tabular}
}
\caption{Experimental results on dependency parsing based on automatic evaluation with StackPTR as the victim model while the Deep Biaffine parser and BiST as Parser B and Parser C.}
\label{tab:parsing-auto-exchange}
\vspace{-0.4cm}
\end{table*}

\section{Related Work}
\paragraph{Attack Design on Un-structured Prediction Model}
Following the success in the image processing area \citep{goodfellow2014explaining}, the idea of adding continuous perturbations to inputs has been applied to tasks in NLP \citep{sato2018interpretable,gong2018adversarial}. 
In order to solve the mapping problem from the modified word vector to the word, \citet{papernot2016crafting} built a special dictionary to select words to replace the original words. 
In addition to replacement manipulation, \citet{samanta2017towards} introduced three modification strategies: removal and addition.
\citet{michel2019evaluation} leveraged atomic character-level operation.
Some attack strategies to generate adversarial examples have been proposed in the sentence level setting. 
\citet{zhao2017generating} searched adversarial examples in the continuous vector space and then used generative adversarial networks \citep{goodfellow2014generative} to map the fixed-length vectors to data instances. 
However, these attackers are only designed for classification tasks or generation tasks and can not be easily applied to structured prediction systems. 

\begin{table}[t]
\centering
\resizebox{0.47\textwidth}{!}{%
\begin{tabular}{r|c|c|c}
\hline
         & \textbf{Generation} & \multicolumn{2}{l}{\textbf{Attacking Success Rate}} \\ \cline{3-4} 
         & \textbf{Fluency} $\uparrow$ & \textit{Token}          & \textit{Sentence}       \\ \hline
AllSame & \textbf{4.19} & 11.4  & 64 \\ \hline
EvalSame     & 3.54  & 13.6        & 62          \\ \hline
Ours     & 3.84   & \textbf{18.3}        & \textbf{72}          \\ \hline
\end{tabular}
}
\caption{Results of human evaluation on different settings of the reference parsers. Higher is better.}
\label{tab:analysis_human}
\vspace{-0.1cm}
\end{table}





\paragraph{Attack Design on Structured Prediction Model}
There is also some prior work on attacking structured prediction models. 
\citet{NIPS2017_7273} proposed to attack structured prediction models in the image processing field, such as those for pose estimation and semantic segmentation.
In a separate line of work, \citet{zugner2019adversarial} proposed to attack graph neural network for node classification. 

\section{Conclusion}
Building an effective adversarial attacker for structured prediction models is challenging. 
The biggest challenge is the sensitivity of the output to small perturbations in the input in structured prediction. 
In this paper, we propose a novel framework to attack structured prediction models in NLP. 
Our framework consists of a structured-output evaluation criterion based on reference models and a seq2seq sentence generator. 
We propose to utilize reinforcement learning to train the sentence generator based on the evaluation criterion. 
Our attack experiments on dependency parsing and POS tagging show that our proposed framework can produce high-quality sentences that can effectively attack current state-of-the-art models. 
Our defense experiments show that adversarial training using the adversarial samples generated by our model can be used to improve the original model.
We believe that our framework is general and can be applied to many other structured prediction tasks in NLP, such as neural machine translation, semantic parsing and so on.

\section*{Acknowledgments}
Kewei Tu and Liwen Zhang are supported by the National Natural Science Foundation of China (61976139).

\bibliographystyle{acl_natbib}
\bibliography{anthology,emnlp2020}

\appendix
\section{Dependency Parsing Experiment Details}
During pretraining, the Deep Biaffine parser and the StackPtr parser is trained by Pytorch 0.4.1, the BiST parser is trained by Dynet. 
\begin{table}[H]
\centering
\begin{tabular}{ll}
\hline
Embedding                      & sskip             \\
Embedding dim                  & 100               \\
Embedding dropout              & 0.33              \\
BiLSTM size                      & 512               \\
BiLSTM depth                     & 3                 \\
BiLSTM dropout                   & 0.33              \\
Arc MLP size                   & 512               \\
Arc MLP dropout                & 0.33              \\
Label MLP size                 & 128               \\
Label MLP dropout              & 0.33              \\
Batch size                     & 32                \\
Optimizer                      & Adam              \\
Learning rate                  & $1e$-$3$          \\ \hline

\end{tabular}%
\caption{Hyper-parameters of pretraining the Deep Biaffine parser. Here sskip is Structured SkipGram \citep{ling2015two}.}
\label{tab:biaffine-parser}
\end{table}

\begin{table}[H]
\centering
\begin{tabular}{ll}
\hline
Embedding                      & sskip             \\
Embedding dim                  & 100               \\
Embedding dropout              & 0.33              \\
BiLSTM size                      & 512               \\
BiLSTM depth                     & 3                 \\
BiLSTM dropout                   & 0.33              \\
Arc MLP size                   & 512               \\
Arc MLP dropout                & 0.33              \\
Label MLP size                 & 128               \\
Label MLP dropout              & 0.33              \\
Batch size                     & 32                \\
Optimizer                      & Adam              \\
Learning rate                  & $1e$-$3$          \\ \hline

\end{tabular}%
\caption{Hyper-parameters of pretraining the StackPtr parser.}
\label{tab:stackptr-parser}
\end{table}

\begin{table}[]
\centering
\begin{tabular}{ll}
\hline
Embedding                      & sskip             \\
Embedding dim                  & 100               \\
POS Embedding dim              & 25                \\
Word Embedding dropout         & 0.25              \\
BiLSTM size                      & 125              \\
BiLSTM depth                     & 2                 \\
MLP size                       & 100               \\
Batch size                     & 32                \\
Window                         & 3                 \\
Optimizer                      & Adam              \\
Learning rate                  & $1e$-$1$          \\ \hline

\end{tabular}%
\caption{Hyper-parameters of pretraining the BiST parser.}
\label{tab:bist-parser}
\end{table}

\begin{table}[H]
\centering
\begin{tabular}{ll}
\hline
Word Embedding                        & sskip                  \\
Word Embedding dim                    & 100                    \\
BiLSTM depth                          & 3                      \\
BiLSTM dim                            & 1024                   \\
Hidden state dropout                  & 0.5                    \\
Optimizer                      & Adam              \\
Learning rate                  & $1e$-$3$          \\
Epoch                                 & 3                      \\ \hline
\end{tabular}%
\caption{Hyper-parameters of pretraining our seq2seq sentence generator for dependency parsing.}
\label{tab:seq2seq-pretrain}
\end{table}

\begin{table}[H]
\centering
\begin{tabular}{ll}
\hline
$\alpha$           & 1     \\
$\beta$            & 0.001   \\
$\gamma$           & 100   \\
UNK weight         & 500   \\
Optimizer          & Adam              \\
Learning rate      & $2e$-$5$          \\
Epoch              & 3                      \\ \hline
\end{tabular}%
\caption{Hyper-parameter of reinforcement training seq2seq sentence generator.
UNK weight is a reward used to control the rate of UNK token. About 6 hours per epoch.}
\end{table}

\paragraph{Retraining the Deep Biaffine parser}
We retrain the parser, all its hyper-parameter is same as the Table~\ref{tab:biaffine-parser} but learning rate is $5e$-$4$. 

\section{POS Tagging Experiment Details}

\begin{table}[H]
\centering
\begin{tabular}{ll}
\hline
\multicolumn{2}{c}{the BiLSTM-CNN-CRF Tagger} \\ \hline
Embedding                      & sskip             \\
Embedding dim                  & 100               \\
Embedding dropout              & 0.33              \\
BiLSTM size                    & 256             \\
BiLSTM depth                   & 1               \\
Label MLP size                 & 256               \\
Label MLP dropout              & 0.5              \\
Bigram                         & True             \\
Batch size                     & 16                \\
Optimizer                      & Adam              \\
Learning rate                  & $1e$-$3$          \\ \hline
\end{tabular}%
\caption{Hyper-parameters during pretraining the BiLSTM-CNN-CRF Tagger.}
\label{tab:tagger}
\end{table}
Reference Tagger:
\begin{itemize}
    \item[-]Stanford POS tagger: \\ \url{http://nlp.stanford.edu/software/stanford-postagger-2015-04-20.zip}
    \item[-] Senna tagger: \\ \url{http://ronan.collobert.com/senna/senna-v3.0.tgz}
\end{itemize}

During pretraining the seq2seq sentence generator, all hyper-parameters are same with Table~\ref{tab:seq2seq-pretrain} but BiLSTM dim is 512.

\begin{table}[H]
\centering
\begin{tabular}{ll}
\hline
$\alpha$              & 1     \\
$\beta$               & 0.001   \\
$\gamma$             & 30   \\
UNK weight         & 0   \\
Optimizer          & Adam              \\
Learning rate      & $5e$-$5$          \\
Epoch              & 3                      \\ \hline
\end{tabular}%
\caption{Hyper-parameter of reinforcement training seq2seq sentence generator. About 22 hours per epoch.}
\end{table}

\paragraph{Retraining the BiLSTM-CNN-CRF Tagger}
We retrain the parser, all its hyper-parameter is same as the Table~\ref{tab:tagger} but learning rate is $1e$-$4$. 

\section{Hyper-Parameter Search}
The criterion used to select all the hyper-parameters is the performance on the development data. We mainly tune the hyper-parameters of the text generator. For example, we choose the dimension of the hidden layer from 20 values in the range of 32 to 2048.
\end{document}